\RequirePackage{fix-cm}
\RequirePackage{amsmath}
\documentclass[smallextended]{svjour3}       % onecolumn (second format)
\smartqed  % flush right qed marks, e.g. at end of proof
\usepackage{graphicx}
\usepackage{xcolor}
\usepackage{amsmath}
\usepackage{amsfonts}
\usepackage{amssymb}
\usepackage{multirow}
\usepackage{eurosym}
\usepackage{url}
\usepackage{cite}
\newcommand{\diag}{\mathop{\mbox{diag}}}

\begin{document}

\title{robROSE: 
A robust approach for dealing with imbalanced data in fraud detection
%\thanks{This work was supported by the BNP Paribas Fortis Chair in Fraud Analytics.}
}
% Grants or other notes about the article that should go on the front
% page should be placed within the \thanks{} command in the title
% (and the %-sign in front of \thanks{} should be deleted)
%
% General acknowledgments should be placed at the end of the article.

%\subtitle{}

\titlerunning{robROSE}        % if too long for running head

\author{Bart Baesens \and Sebastiaan H\"oppner \and Irene Ortner \and Tim Verdonck
}


\institute{Bart Baesens \at 
            Faculty of Ecomomics and Business, KU Leuven\\
            Naamsestraat 69, 3000 Leuven, Belgium\\
            bart.baesens@kuleuven.be
            \and
            Sebastiaan H\"oppner \at
            Department of Mathematics, KU Leuven\\
            Celestijnenlaan 200B, 3001 Leuven, Belgium\\
            sebastiaan.hoppner@kuleuven.be
            \and
            Irene Ortner \at
            Applied Statistics GmbH\\
            Taubstummengasse 4/10, 1040 Vienna, Austria \\
            irene.ortner@applied-statistics.at
            \and
            Tim Verdonck \at
            Department of Mathematics, University of Antwerp\\
            Middelheimlaan 1, 2020 Antwerp, Belgium\\
            tim.verdonck@uantwerp.be
}

\date{Received: date / Accepted: date}

\maketitle

\begin{abstract}
A major challenge when trying to detect fraud is that the fraudulent activities form a minority class which make up a very small proportion of the data set.  In most data sets, fraud occurs in typically less than $0.5\%$ of the cases. Detecting fraud in such a highly imbalanced data set typically leads to predictions that favor the majority group, causing fraud to remain undetected. We discuss some popular oversampling techniques that solve the problem of imbalanced data by creating synthetic samples that mimic the minority class. A frequent problem when analyzing real data is the presence of anomalies or outliers. When such atypical observations are present in the data, most oversampling techniques are prone to create synthetic samples that distort the detection algorithm and spoil the resulting analysis. A useful tool for anomaly detection is robust statistics, which aims to find the outliers by first fitting the majority of the data and then flagging data observations that deviate from it. In this paper, we present a robust version of ROSE, called robROSE, which combines several promising approaches to cope simultaneously with the problem of imbalanced data and the presence of outliers. The proposed method achieves to enhance the presence of the fraud cases while ignoring anomalies. The good performance of our new sampling technique is illustrated on simulated and real data sets and it is shown that robROSE can provide better insight in the structure of the data. The source code of the robROSE algorithm is made freely available.

\end{abstract}

\section{Introduction}

%{\color{gray}
%Note differences to ROSE paper
%\begin{itemize}
%    \item ROSE: Kernel estimator with assumption of uncorrelated features.
%    \item Our sampling strategy: we include artificial data only for training and also use real data. To test a model it is essential to look at it's performance for independent real data.
%    \item When we apply ROSE algorithm in this paper we follow our strategy to include real observations and not use artificial observations for testing.
%\end{itemize}
%}

The Association of Certified Fraud Examiners (ACFE) estimates that a typical organization loses $5\%$ of its revenues to fraud each year. The Nilson Report, a publication covering news and analysis of the global payment industry, reported that global card fraud losses equaled $22.8$ billion in 2016, which is an increase of 4.4 percent over 2015. According to the European Insurance Committee, fraud takes up $5$ to $10\%$ of the claim amounts paid for non-life insurance and the FBI estimates that the total cost of non-health insurance fraud in the US is more than $\$40$ billion per year. The national audit, tax and advisory firm Crowe Clark Whitehill, together with the University of Portsmouth’s Centre for Counter Fraud Studies (CCFS), estimate that the fraud epidemic costs the UK $\pounds 110$ billion a year. According to their report, businesses lose an average of $6.8\%$ of total expenditure and it is concluded that fraud is the last great unreduced business cost.
%https://www.acfe.com/press-release.aspx?id=4294973129
%https://losspreventionmedia.com/credit-card-fraud-statistics-and-insights/
% https://nilsonreport.com
%https://www.fbi.gov/stats-services/publications/insurance-fraud
%https://www.globalbankingandfinance.com/fraud-epidemic-costs-the-uk-110-billion-and-3-2-trillion-globally/
These are just a few numbers to indicate the severity of the fraud problem. We can also see that losses due to fraudulent activities keep increasing each year and affect organizations worldwide. Therefore, fraud detection and prevention is more important than ever before and developing powerful fraud detection systems is of crucial importance in order to reduce losses.

The Oxford Dictionary defines fraud as ``wrongful or criminal deception intended to result in financial or personal gain''. This definition captures the essence of fraud but it does not very precisely describe the nature and characteristics of fraud. A more thorough and detailed characterization of the multifaceted phenomenon of fraud is provided by \cite{van2016gotcha}: ``fraud is an uncommon, well-considered, imperceptibly concealed, time-evolving and often carefully organized crime which appears in many types of forms''. This definition highlights five characteristics that are associated with particular challenges related to developing a fraud detection system and as such also describes the requirements of a successful fraud detection system.

In this paper, we will focus on the first emphasized characteristic and associated challenge of this detailed definition, namely the fact that fraud is uncommon or rare. For example, in the setting of credit card fraud, typically less than $0.5\%$ of transactions are fraudulent. Such a problem is commonly referred to as the needle in a haystack problem. Independent of the exact setting or application, only a minority of the involved population of cases typically concerns fraud, of which even a more limited number will be known to actually concern fraud. Highly imbalanced or skewed data make it difficult to detect fraud, since the fraudulent cases are covered by the non-fraudulent ones, and to learn from historical cases to build a powerful fraud detection system since only few examples are available. 

A stream of literature has reported upon the adoption of data-driven approaches for developing fraud detection systems, see for example \cite{phua2010comprehensive} and \cite{ngai2011application}. These data-driven methods have three important benefits towards an expert-based approach: they significantly improve the efficiency of fraud detection systems, they are more objective and they are easier to maintain. From a machine learning perspective, the task of detecting fraudulent transactions is a binary classification problem. Popular data-driven techniques for fraud detection are logistic regression and  decision trees among others. These methods are so-called white box models and therefore yield a clear explanation behind how they reach their classification. These models hence enable the user to understand the underlying reasons why the model signals an observation to be suspicious. Besides their interpretability, they are also operationally efficient. To increase the detection power, these simple analytical models can be extended by adding a penalty or regularization term, or using the idea of ensemble learning. Alternative complex techniques are neural networks and support vector machines. However, they suffer from a very important drawback which is not desirable from a fraud detection perspective: they are black box models which means that they are very difficult to interpret. 

It is not our aim to give a detailed overview of the various machine learning techniques that could be applied for fraud detection. Instead, we focus on the imbalance or skewness of the data, meaning that typically there are plenty of historical examples of non-fraudulent cases, but only a limited number of fraudulent cases. This problem typically causes an analytical technique to experience difficulties in learning to create an accurate model. Most classifiers faced with a skewed data set typically tend to favor the majority class. In other words, the classifier tends to label all observations as non-fraudulent since it then already achieves a classification accuracy of more than $99\%$. Classifiers typically learn better from a more balanced distribution. Two ways to accomplish this is by random undersampling, whereby non-fraudulent transactions in the training set are removed, or random oversampling, whereby fraudulent transactions in the training set are replicated. Better results are obtained when applying synthetic oversampling, which oversample the minority class by creating synthetic examples to improve the performance of the fraud detection model. Synthetic Minority Oversampling Technique (SMOTE) is the first and probably the most well-known synthetic oversampling technique to deal with skewed data \cite{chawla2002smote}. Since then, many variants and alternatives for SMOTE have been presented in the academic literature. In this paper we focus on ROSE \cite{menardi2014rose}, which generates new minority samples based on the kernel density estimate around existing, real minority cases. When outliers or anomalies are present in the data, ROSE unfortunately also creates synthetic examples based on these outliers. This may distort the detection algorithm and actually lower the performance. Therefore, we introduce a robust version of the ROSE algorithm, which is not sensitive to the presence of outliers. Synthetic examples  are then only generated from observations which are considered clean or normal observations. Moreover, our robust version also takes the covariance structure of the data into consideration to create more realistic artificial observations. 

The remainder of the article is organized as follows. In Section 2, we will present SMOTE and other popular sampling techniques to solve the class imbalance problem. Section 3 describes performance measures for classification that are suited to evaluate a fraud detection model. In section 4 our robust oversampling technique for dealing with imbalanced data in fraud detection is proposed. We illustrate its good performance on simulated data in Section 5, whereas Section 6 analyzes a credit card transaction dataset for fraud. Section 7 shows that the proposed methodology can also be used in other domains than fraud detection and Section 8 concludes.

\section{Selection of popular sampling techniques}

The literature review given by \cite{he2009learning} shows the large extent to which techniques for handling imbalanced learning problems are researched. The solutions to imbalanced data sets can be divided into four categories: sampling-based methods, cost-based methods, kernel-based methods and active learning-based methods \cite{he2009learning}. In this paper, we are interested in sampling methods and provide a brief overview of the works performed in this category. A summary of the work performed in the other categories can be found in \cite{he2009learning}. For more information about cost-sensitive approaches in the context of fraud detection, we refer to \cite{hand2008performance,bahnsen2013cost}. \cite{zhu2019iric} recently implemented techniques for imbalanced classification in an R library called IRIC.

Sampling methods operate at the data level as they change the distribution between the majority class and the minority class samples of the imbalanced data set. The balanced data set is then provided to the classification algorithm which typically learns better from a balanced distribution than from an imbalanced one and so the detection rate of minority cases is improved \cite{weiss2001effect}. Balancing the distribution can be done by either reducing the majority class samples or by adding minority class samples. The former is called undersampling and the latter is called oversampling. Random undersampling is the simplest form of undersampling as it randomly takes away samples from the majority class, while informed undersampling uses some statistical knowledge to remove majority samples \cite{liu2008exploratory}.

Our focus, however, lies in oversampling methods. Random oversampling simply duplicates samples from the minority class which could lead to very specific rules and hence overfitting \cite{holte1989concept}. Synthetic oversampling, on the other hand, adds new information to the original data set by generating synthetic minority class samples to improve the performance of the classifier. Various synthetic oversampling methods exist in the literature such as Synthetic Minority Oversampling TEchnique (SMOTE) \cite{chawla2002smote}, Borderline-SMOTE \cite{han2005borderline}, Adaptive Synthetic Sampling Technique (ADASYN) \cite{he2008adasyn}, MWMOTE \cite{barua2012mwmote} and ROSE \cite{menardi2014rose}. The ways in which these oversampling methods generate synthetic minority samples can be described based on how they answer the following three questions.
\begin{enumerate}
    \item Which minority samples do we want to oversample? All of them or are there minority outcasts (i.e. ``anomalous minorities'') which we exclude from the process?
    \item By how much do we want to oversample the minority cases? Should we oversample some minority cases more than others?
    \item How should we oversample or, in other words, how do we create synthetic minority cases? % On a line segment between two (minority) cases or maybe in a region around a minority case? How many neighbors around a minority case should we choose from, fixed (e.g. k = 5) or adaptive?
\end{enumerate}
SMOTE \cite{chawla2002smote} considers all minority samples and creates a synthetic case $z$ as a point on the line segment between two minority cases $x$ and $y$:
\[z = x + \alpha(y-x)\]
where $\alpha$ is a random number in the unit interval $[0, 1]$. Borderline-SMOTE \cite{han2005borderline} does not deal with every minority class sample, but instead it identifies so-called border-line minority class samples which are most likely to be misclassified by a classifier. These border-line samples are then used for generating the synthetic samples in the same way as SMOTE does. ADASYN \cite{he2008adasyn} assigns weights to the minority class samples, so minority samples are not treated equally. A large weight helps in generating many synthetic samples from the corresponding minority class sample. %According to \cite{barua2012mwmote}, the previously mentioned methods may generate the wrong synthetic minority samples in some scenarios and make learning tasks harder.
MWMOTE \cite{barua2012mwmote} first identifies the hard-to-learn informative minority class samples and assigns them weights according to their euclidean distance from the nearest majority class samples. Next, it generates the synthetic samples from the weighted informative minority class samples in the same way as SMOTE using a clustering approach. This is done in such a way that all the generated samples lie inside some minority class cluster. All previously mentioned methods are related to SMOTE in the sense that they create synhetic minority samples on a line segment between two existing minority cases. ROSE \cite{menardi2014rose}, on the other hand, generates new minority samples based on the kernel density estimate around existing, real minority cases. 

A practical question concerns the optimal, non-fraud/fraud rate, which should be the goal by doing oversampling. One popular trial-and-error approach to determine the optimal class distribution works as follows: In the first step, a classifier is built on the original data set with the skewed class distribution, which has, for example, $99\%$ non-fraudsters and $1\%$ fraudsters. The performance of this model is then measured on an independent validation data set. In a next step, oversampling is used to change the class distribution to for example, $90\%$ and $10\%$. Again, the model is evaluated. Subsequent models are built on samples of 85\% versus 15\%, 80\% versus 20\%, 75\% versus 25\%, and so on. Each time the performance is recorded. When the performance starts to stagnate or drop, the procedure stops and the optimal odds ratio is found. Although it does depend on the data characteristics and quality, practical experience shows that the ratio 90\% non-fraudsters versus 10\% fraudsters is quite commonly used in the industry. In the next section, we review performance measures that are suited for imbalanced data sets.

%\subsection{Adjusting Posterior Probability Estimates}
%Recall that the key idea of undersampling, oversampling, SMOTE and ROSE is to adjust the class priors to enable the analytical technique to create a meaningful model that discriminates between two classes. By doing so, the class posteriors become biased. This is no problem if the modeler is interested only in ranking the observations in terms of their probability. However, if well-calibrated probabilities are needed, then the posterior probabilities need to be adjusted. One straightforward way to do this is by using the following formula~\citep{saerens2002adjusting}:
%$$p(C_i|x) = \frac{ \frac{p(C_i}{p_r(C_i)} p_r(C_i) | x) }{ \sum^2_{j=1} \frac{p(C_j}{p_r(C_j)} p_r(C_j) | x)  }$$
%where $C_i$ represents class $i$, $p(C_i)$ the prior probabilities (e.g. 1\% and 99\%), $p_r(C_i)$ the resampled prior probabilities due to oversampling, undersampling, SMOTE or ROSE (e.g. 50\% and 50\%), and $p_r(C_i|x)$ the posterior probabilities for observation $x$ to belong to class $C_i$, as given by the predictive model using the resampled data.

\section{Model evaluation for imbalanced data sets}

When performing a classification task, assessing the classifier's quality plays a crucial role that is at least as important as estimating the model, especially in a class imbalance context. By labeling one class as a positive and the other class as a negative, the performance of binary classification algorithms is typically measured by using a confusion matrix as illustrated in Table 1. The rows represent the class as predicted by the model and the columns are the actual class. Typically, the minority class (i.e. fraud) is used as the positive class and the majority class (i.e. legitimate) as the negative class. In the confusion matrix, we count the following numbers:
\begin{itemize}
    \item TN $=$ number of correctly classified negative cases (True Negatives), e.g. correctly identified legitimate cases
    \item FP $=$ number of negative cases incorrectly classified as positive (False Positives), e.g. legitimate cases wrongly labeled as fraudulent
    \item FN $=$ number of positive cases incorrectly classified as negative (False Negatives), e.g. undetected fraud cases
    \item TP $=$ number of correctly classified positive cases (True Positives), e.g. detected fraud cases
\end{itemize}
\begin{table}[b!]
\centering
\begin{tabular}{c cc}
\hline
 & Actual  & Actual  \\
 & negative  & positive  \\\hline
 Predicted & True negatives  & False negatives  \\
 negative & (TN) & (FN) \\
  & & \\
 Predicted & False positives &  True positives \\
 positive & (FP) &  (TP) \\ \hline
\end{tabular}
\caption{Confusion matrix for binary classification.}
\label{tab:confusion_matrix}
\end{table}
Several performance measures can be derived from Table \ref{tab:confusion_matrix}. The two most common are predictive accuracy and error rate which are defined as
\[Accuracy = \frac{TP + TN}{TP + FP + TN + FN}\text{ and }Error\text{ }rate = 1 - Accuracy.\]
The main problem associated with the accuracy and error rate measures is their dependence on the distribution of positive class and negative class samples in the data set.  This makes them not well suited for imbalanced learning problems \cite{he2009learning}.
Other evaluation metrics can be derived from Table \ref{tab:confusion_matrix} to assess learning from imbalanced data, such as:
\begin{align*}
    Precision &= \frac{TP}{TP + FP} \\
    Recall &= \frac{TP}{TP+FN}\\
    F_1\text{-}measure &= \frac{2\cdot precision \cdot recall}{recall + precision}
\end{align*}
Precision is the proportion of actual fraud cases among the cases that are predicted as fraud by the model. Recall or sensitivity (true positive rate), on the other hand, is the proportion of fraudulent transfers that are detected by the model. The $F_1$-measure combines precision and recall as an harmonic mean for maximizing the performance on a single class. Hence, it is used for measuring the performance of the classifier on the minority class samples.
\begin{figure}[htbp]
	\centering
	\includegraphics[width=0.95\textwidth]{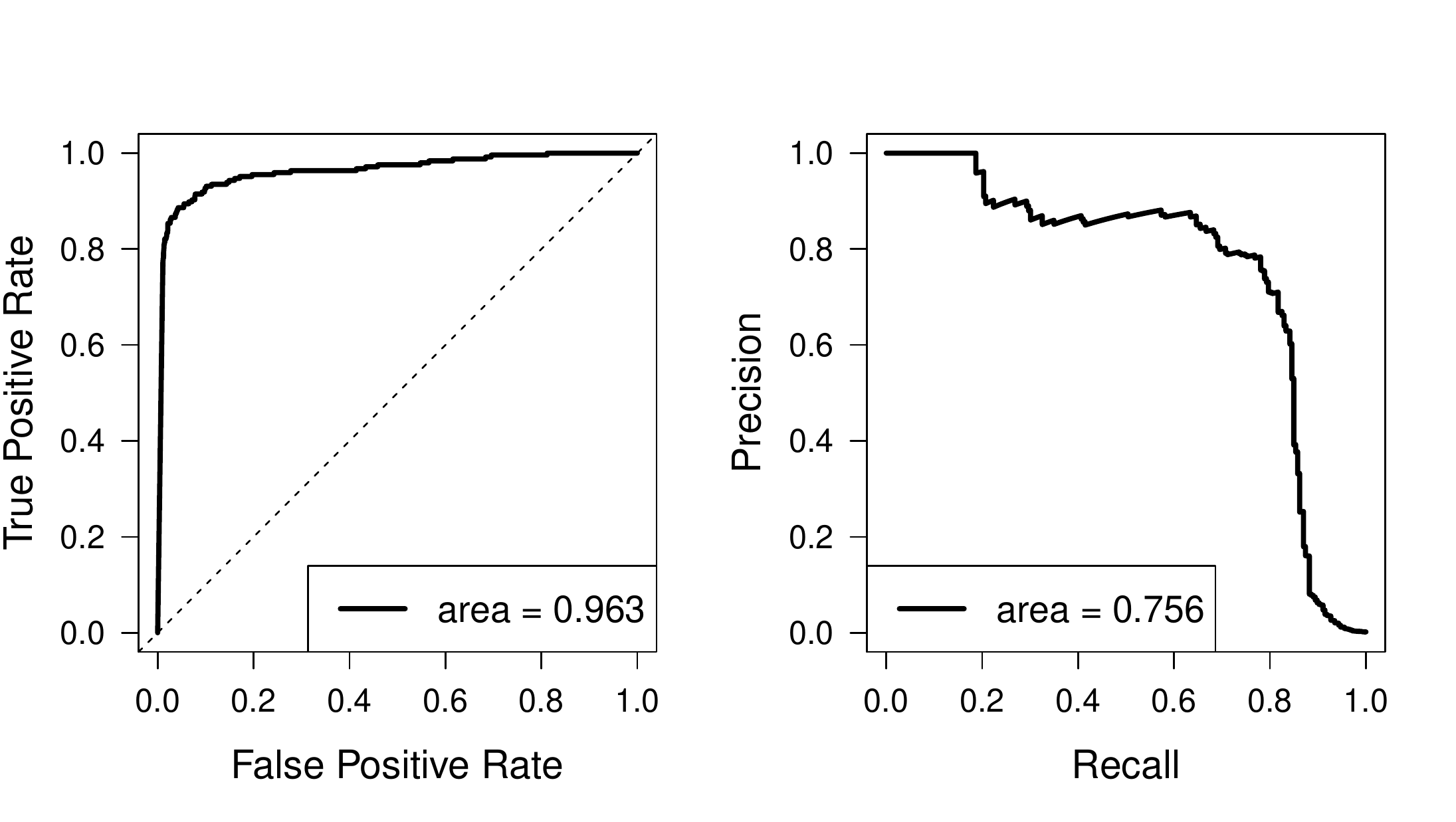}
	\caption{(Left) example of an ROC curve. (Right) example of an Precision-Recall curve.}
	\label{fig:ROC_PRcurve}
\end{figure}
Note that the measures above depend on the cut-off value.

Each of the measures above are calculated for a given confusion table that is based on a certain cutoff value. 
The receiver operating characteristic (ROC) curve, as shown on the left plot in Figure \ref{fig:ROC_PRcurve}, is obtained by plotting for each possible cutoff value the false positive rate ($FPR$) on the $X$-axis and the sensitivity or true positive rate ($TPR$) on the $Y$-axis, where
\[FPR = \frac{FP}{FP + TN} \quad\text{and}\quad TPR = \frac{TP}{TP + FN}.\]
The false positive rate is also referred to as the inverted specificity (i.e. 1-specificity) where specificity or true negative rate is the total number of true negatives divided by the sum of the number of true negatives and false positives. \cite{provost1998t} argued that ROC curves as an alternative to accuracy estimation for comparing classifiers would enable stronger and more general conclusions. For more information about ROC curves we refer to \cite{krzanowski2009roc} and \cite{swets2014signal}. 
% Use of this method dates back to World War II when the ability of radar operators (receivers) was tested to determine whether a blip on the radar screen represented an object (signal, a TP result) or noise (a FP result), hence, the name. 
 %Altman DG, Bland JM. Diagnostic tests 3: receiver operating characteristic plots. BMJ. 1994;309:188. 10.1136/bmj.309.6948.188 [PMC free article] [PubMed] [CrossRef] [Google Scholar]
%Metz CE. Basic principles of ROC analysis. Semin Nucl Med. 1978;8:283–98. 10.1016/S0001-2998(78)80014-2 [PubMed] [CrossRef] [Google Scholar]
%Swets JA. Measuring the accuracy of diagnostic systems. Science. 1988;240:1285–93. 10.1126/science.3287615 [PubMed] [CrossRef] [Google Scholar]

Probably the most popular tool today to measure the performance of a classifier is then the area under this ROC curve \cite{fawcett2004roc, fawcett2006introduction, ling2003auc}, usually known as AUC. 
%Unlike the $F_1$-measure, AUC is not sensitive to the distribution between the positive class and negative class samples, thus suitable for performance comparison of different classifiers \cite{fawcett2004roc, fawcett2006introduction}. 
The AUC of a classifier can be interpreted as being the probability that a randomly chosen minority case (i.e. fraud) is predicted a higher score than a randomly chosen majority case (i.e. legitmate). Therefore, a higher AUC indicates superior classification performance.

However, when dealing with highly imbalanced data sets, AUC  (and ROC curves) may be too optimistic and the area under the Precision-Recall curve (AUPRC) gives a more informative picture of an algorithm's performance \cite{davis2006relationship}. 
As the name suggest, the Precision-Recall curve (right plot in Figure \ref{fig:ROC_PRcurve}) plots the precision (Y-axis) against the recall (X-axis).
Both ROC as PR curves use the recall or sensitivity, but the ROC curve also plots the $FPR$ whereas PR curves focus on precision. In the denominator of $FPR$, one sums the number of true negatives and false positives. In highly imbalanced data, the number of negatives (good observations) is much larger than the number of positives (fraudulent observations) and hence the number of true negatives is typically very high compared to the number of false positives. Therefore, a large increase or decrease in the number of false positives will have almost no impact on $FPR$ in the ROC curves. Precision, on the other hand, compares the number of false positives to the number of true positives and hence copes better with the imbalance between positives and negatives. Because Precision is more sensitive to class imbalance, the area under the Precision-Recall curve is better to highlight differences between models for highly imbalanced data sets. 

%An alternative approach (outside the scope of this paper) to handle imbalanced data is to use cost curves as recommended in \cite{drummond2000explicitly,drummond2004roc}. 

\section{Methodology: robROSE}

We introduce a robust version of the ROSE algorithm, which does not oversample minority outcasts and additionally takes the covariance structure of the data into consideration. Artificial observations are generated only from observations which are considered clean or normal observations. Anomalous observations can have a huge influence on the model estimation. To reduce the influence of such observations robust modelling techniques can be used. However, even with robust models we may introduce too many clustered outliers with oversampling such that the model is distorted. Therefore, we introduce a method, called robROSE, as a robust alternative to the ROSE algorithm which uses only non-outlying minority cases for oversampling.

For the identification of outliers in the minority group we use robust Mahalanobis distances, i.e. Mahalanobis distances with respect to the robust center $\hat{\boldsymbol{\mu}}_1$ and scatter estimator $\hat{\boldsymbol{\Sigma}}_1$ of the minority samples. More precisely, we apply the fast MCD estimator \cite{rousseeuw1999fast} on the minority samples. Since we compute the covariance structure for outlier identification we can also use this information to define a density around each observation of the minority samples to generate artificial observations. In contrast to the ROSE algorithm this offers the advantage that we are able to consider the covariance structure of the minority samples when we generate artificial observations.

The algorithm is outlined in detail in Algorithm 1. Let $\boldsymbol{X}\in\mathbb{R}^{n\times p}$ denote the full data set, $\boldsymbol{X}_1$ and $\boldsymbol{X}_0$ denote its subsets belonging to the minority class and majority class, respectively. We compute the center and scatter of the minority samples with the fast MCD estimator and obtain $\hat{\boldsymbol{\mu}}_1$ and $\hat{\boldsymbol{\Sigma}}_1$, respectively. Then we compute robust Mahalanobis distances (MD) for each $\boldsymbol{x}_i\in\boldsymbol{X}_1$ to identify outliers within the group of minority samples:
\begin{equation*}
    MD(\boldsymbol{x}_i, \hat{\boldsymbol{\mu}}_1, \hat{\boldsymbol{\Sigma}}_1) = \sqrt{ (\boldsymbol{x}_i - \hat{\boldsymbol{\mu}}_1)^T\hat{\boldsymbol{\Sigma}}_1^{-1}(\boldsymbol{x}_i - \hat{\boldsymbol{\mu}}_1)}
\end{equation*}

The squared Mahalanobis distances follow a $\chi^2$-distribution with $p$ degrees of freedom. Observations with squared MD larger than the 99.9\% quantile of the $\chi^2$-distribution are considered as outliers and are excluded from oversampling. The cutoff quantile of 99.9\% is relatively large compared to other applications of outlier detection. We want to avoid loosing good observations and therefore we take this rather conservative approach to outlier detection since the number of observations in the minority group is already rather small in the context of imbalance classification. Note that the covariance estimation is based on a MCD estimator with 50\% breakdown point.

After identifying the collection of non-outlying minority samples, this set of observations is used for oversampling. We start by randomly selecting one of the non-outlying minority samples denoted by $\boldsymbol{x}_j$. An artificial observation is then generated from the multivariate normal distribution with center $\boldsymbol{x}_j$ and covariance matrix
\begin{equation}
    \hat{\boldsymbol{\Sigma}}_{x} = \diag(H,...,H)\hat{\boldsymbol{\Sigma}}_1\diag(H,...,H)
\end{equation}
with $H=h*c$ and $c=(4/((p+2)n))^{(1/(p+4))}$ and a tuning constant here set to $h=0.5$. 
$\hat{\boldsymbol{\Sigma}}_{x}$ is a shrunken version of $\hat{\boldsymbol{\Sigma}}_1$ with the same shape, but different size. Together with $\boldsymbol{x}_j$ as center, $\hat{\boldsymbol{\Sigma}}_{x}$ defines a multivariate normal distribution describing the neighbourhood of $\boldsymbol{x}_j$. Random samples drawn from this distribution are similar to $\boldsymbol{x}_j$.

% Alternative: http://www.stat.rice.edu/~scottdw/ss.nh.pdf on multivariate Kernel estimation and rule of thumbs for c
The motivation for $c=(4/((p+2)n))^{(1/(p+4))}$ originates from Gaussian kernels with diagonal smoothing matrix as it is used for ROSE \cite{bowman1997applied}. We follow this approach such that our proposed method is equivalent to ROSE in case of a diagonal covariance matrix and no identified outliers.

\noindent\makebox[\linewidth]{\rule{\textwidth}{0.4pt}}
\textbf{ Algorithm 1:} robROSE.

\vspace{-2mm}
\noindent\makebox[\linewidth]{\rule{\textwidth}{0.4pt}}
Let $\boldsymbol{X}\in\mathbb{R}^{n\times p}$ denote the full data set and let $\boldsymbol{X}_1$ and $\boldsymbol{X}_0$ denote its subset belonging to the minority class and majority class, respectively.
%\vspace{-2mm}
\begin{itemize}
\item Calculate robust covariance $\hat{\boldsymbol{\Sigma}}_1$ and robust center $\hat{\boldsymbol{\mu}}_1$ of $\boldsymbol{X}_1$ with the MCD estimator.
\item Calculate robust Mahalanobis distance for $\boldsymbol{X}_1$
\begin{equation*}
    MD(\boldsymbol{x}_i, \hat{\boldsymbol{\mu}}_1, \hat{\boldsymbol{\Sigma}}_1)^2 = (\boldsymbol{x}_i - \hat{\boldsymbol{\mu}}_1)^T\hat{\boldsymbol{\Sigma}}_1^{-1}(\boldsymbol{x}_i - \hat{\boldsymbol{\mu}}_1)
\end{equation*}
\item Identify the index set of non-outlying minority observations
\begin{equation}
    I=\{i : \boldsymbol{x}_i \in \boldsymbol{X}_1 \text{ and } MD(\boldsymbol{x}_i)^2 < \chi^2_{p,0.999}\}
\end{equation}
with $\chi^2_{p,0.999}$ the 99.9\% quantile of the $\chi^2$-distribution with $p$ degrees of freedom.
\item Generate artificial observations until the desired balance is reached by repeating the following:
\begin{itemize}
\item Randomly select the index $j\in I$ and obtain the observation $\boldsymbol{x}_j$
\item Generate a random sample $\boldsymbol{z}\sim N_p\left(\boldsymbol{x}_j, \hat{\boldsymbol{\Sigma}}_{x} \right)$ from the multivariate normal distribution with center $\boldsymbol{x}_j$ and scatter matrix
\begin{equation*}
    \hat{\boldsymbol{\Sigma}}_{x} = \diag(H,...,H)\hat{\boldsymbol{\Sigma}}_1\diag(H,...,H)
\end{equation*}
where $H=h*c$ and $c=(4/((p+2)n))^{(1/(p+4))}$ and $h=0.5$.
\end{itemize}
\item Return a matrix $\boldsymbol{Z}$ of artificial observations.
\end{itemize}

\noindent\makebox[\linewidth]{\rule{\textwidth}{0.4pt}}

This oversampling technique can also be applied if categorical variables are present in which case the artificial sample inherits the same categories as the minority sample on which it is based.

The \texttt{R} code implementation of robROSE is available in the \texttt{R} package \texttt{robROSE} at github.com/SebastiaanHoppner/robROSE.%\newline
%\text{}

%\subsection{Oversampling and model estimation}
Oversampling can be considered as a preprocessing step for the training data which is independent of the chosen model class. The artificial observations generated from the training data are joined with the real observations of the training data. The resulting data set is then used for model estimation.

In order to be able to assess the model it is crucial to have an independent test data set. In \cite{menardi2014rose} it is recommended to use artificial data for the model assessment as well. They argue that it can help improving the estimation of the posterior probabilities of the  minority class. However, if tuning constants are not properly chosen, then this way of assessing the model's performance can not only result in bad models but also in misleading evaluation. Therefore, we highly recommend to use part of the original data as test data for the model evaluation.

%\section{Illustration}

\begin{figure}[b!]
	\centering
	\begin{minipage}[b]{0.48\textwidth}
	\centering
		\includegraphics[width=0.90\textwidth]{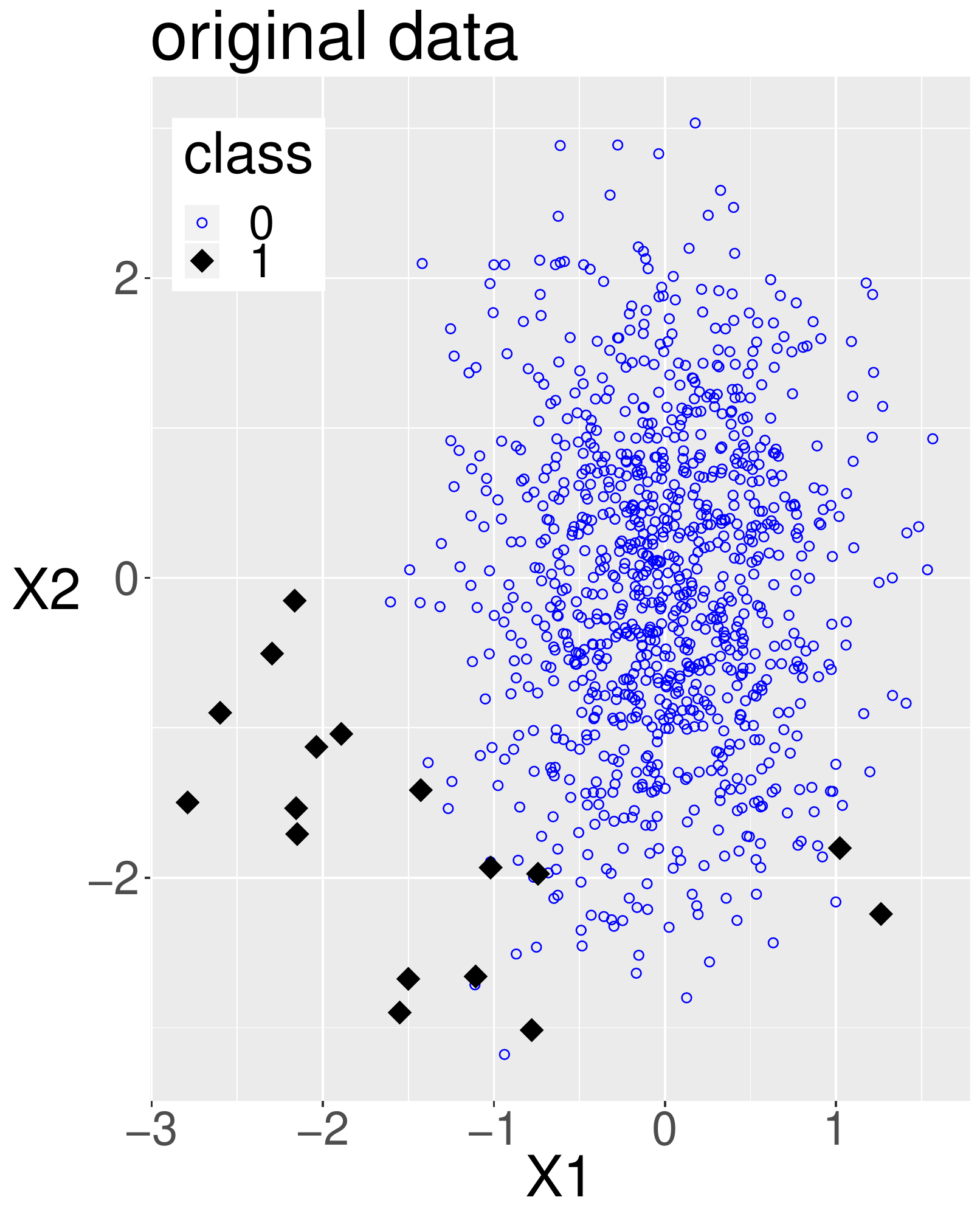}
	\end{minipage}
	\hfill
	\begin{minipage}[b]{0.48\textwidth}
	\centering
		\includegraphics[width=0.90\textwidth]{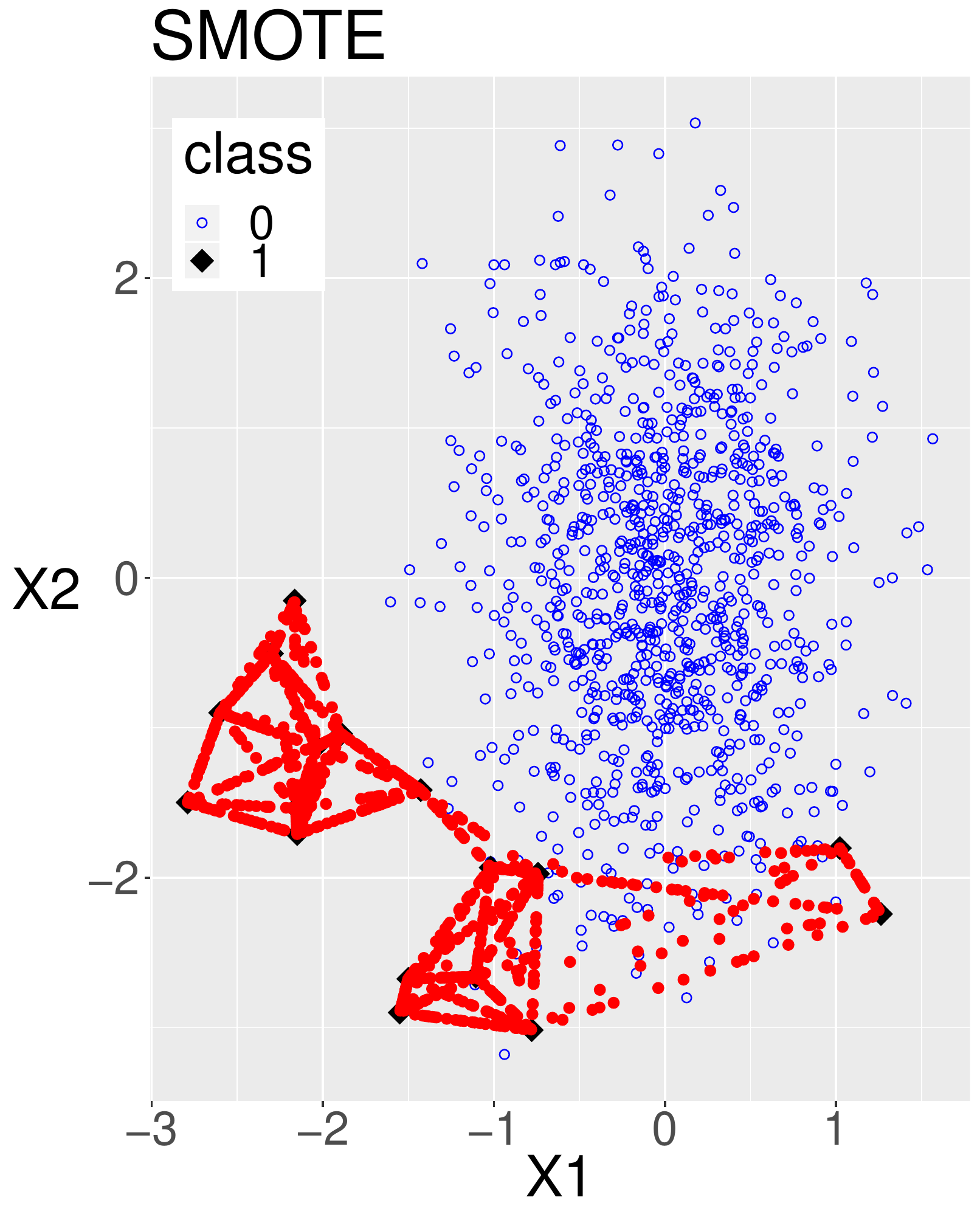}
	\end{minipage}
   \begin{minipage}[b]{0.48\textwidth}
   \centering
	   \includegraphics[width=0.90\textwidth]{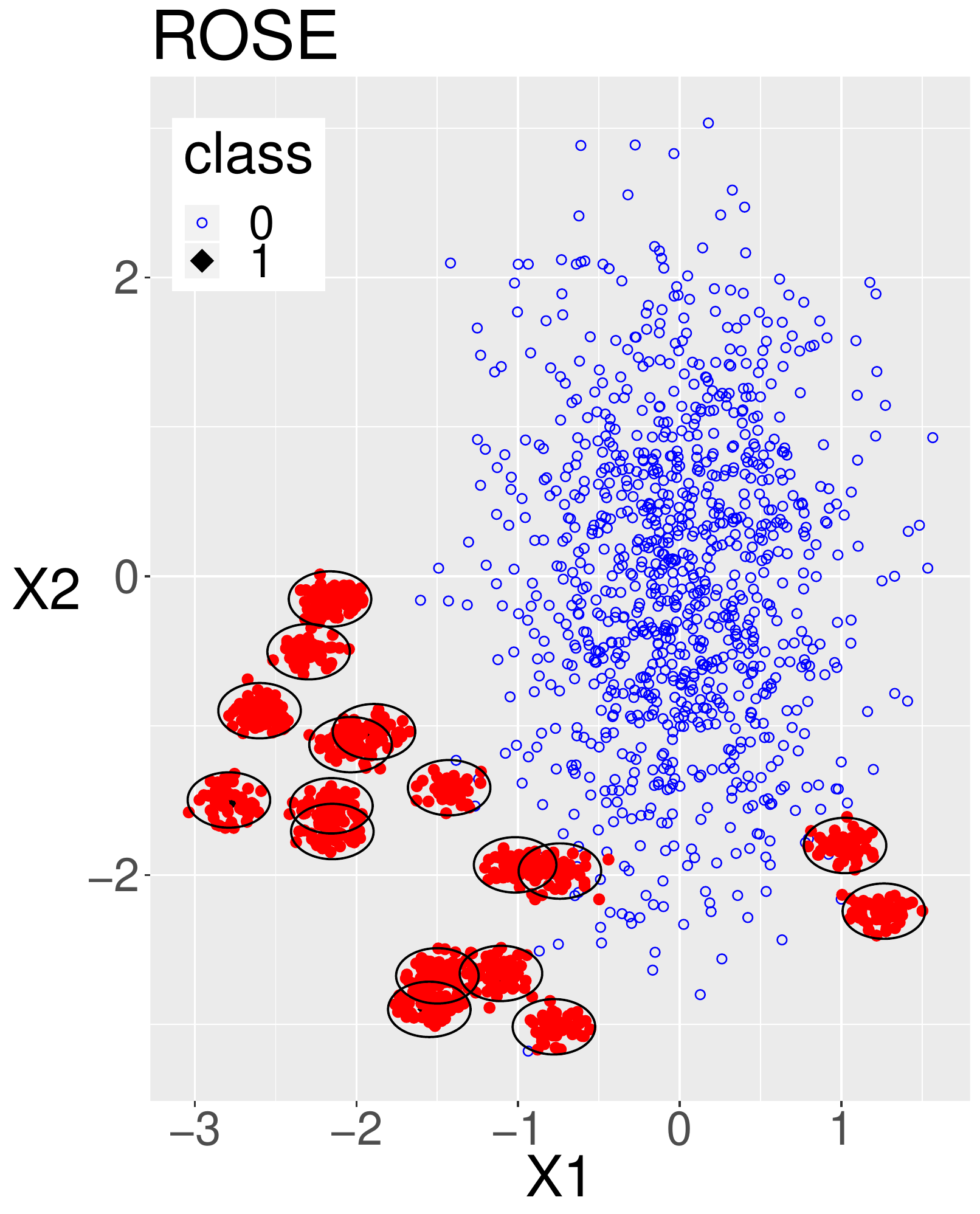}
   \end{minipage}
    \hfill
   \begin{minipage}[b]{0.48\textwidth}
   \centering
	  \includegraphics[width=0.90\textwidth]{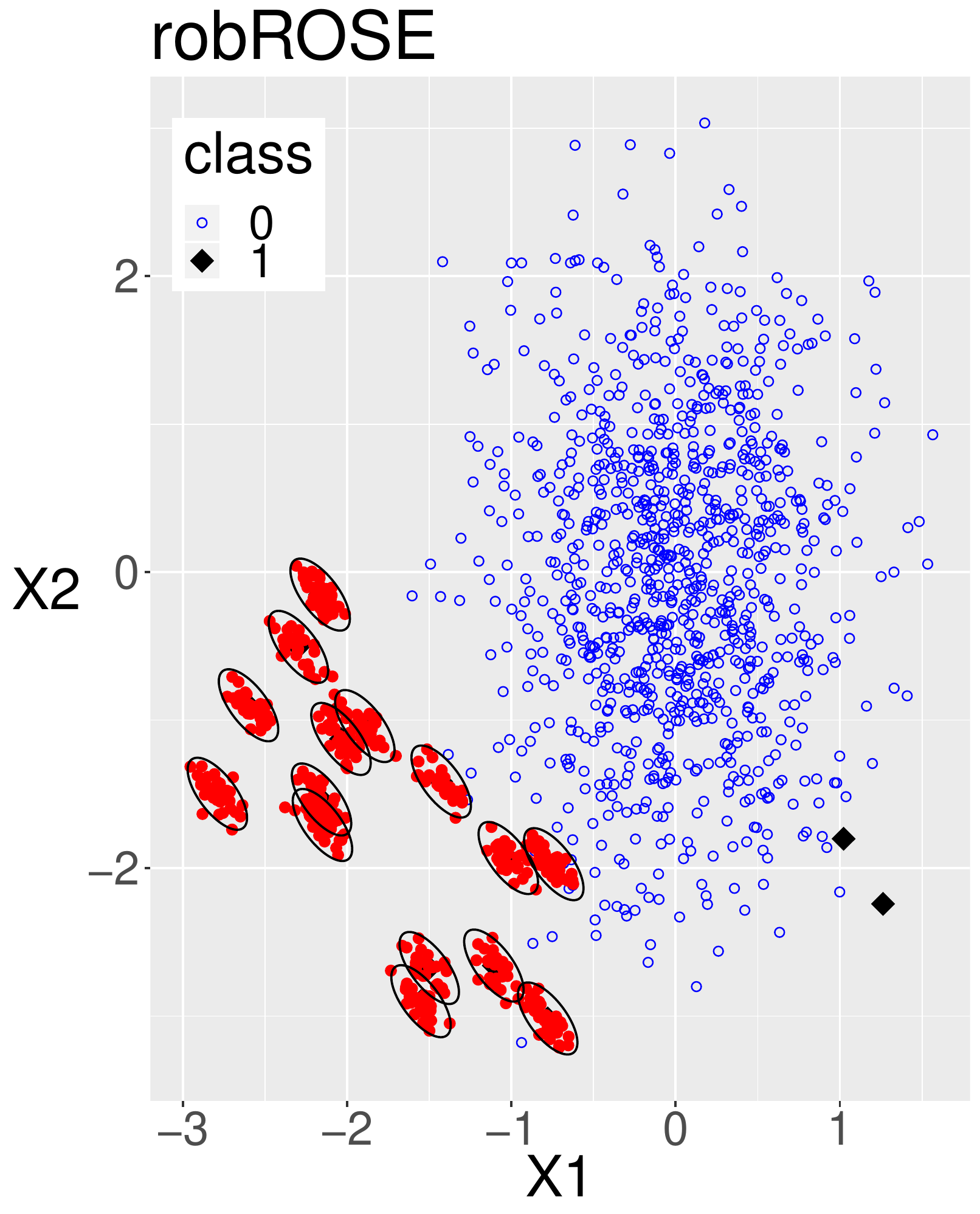}
   \end{minipage}
   	\caption{Illustration of SMOTE, ROSE and robROSE on small artificial dataset.}
	\label{fig:illustration_oversampling}
\end{figure}

We will illustrate the oversampling techniques SMOTE, ROSE and our robROSE on an artificial toy example.
Consider the data in the top-left corner of Figure \ref{fig:illustration_oversampling} which contains both legitimate cases (class 0 in blue) and fraud cases (class 1 in black). Notice how the two fraud cases on the right hand side deviate from the other fraud cases. In order to balance the distribution between the two classes, SMOTE creates synthetic minority (i.e. fraud) samples (in red). However, many of these synthetic cases coincide with regular cases due to the two outlying fraud cases. This further complicates analytical models in detecting fraud. Similarly, ROSE creates synthetic cases around the neigborhood of the two outlying minority samples. Our proposed robust version, robROSE, does not oversample the minority outcasts and additionally takes the covariance structure of the data into consideration as illustrated by the elliptical contours.
%``... while the new examples generated by ROSE lie in the elliptical neighborhood of the observed rare data, synthetic examples generated by means of SMOTE lie along the line segments joining the minority class examples. Thus, the use of SMOTE risks the decision region associated to the rare class in the features space not to be enlarged enough. ...''

\section{Simulation study}
\subsection{Simulation setting}

\subsubsection{Data simulation}

A simulation study is performed to study the properties of the proposed method. We use a similar setting as in \cite{menardi2014rose}. For each class, we generate data from a multivariate normal distribution. The covariance matrix used for the majority class is a diagonal matrix with ones in the diagonal, denoted by $\Sigma_0$. For the minority class, the covariance matrix $\Sigma_1$ has ones in the diagonal, 0.5 in the first off-diagonals and zeros elsewhere. Let $\mu_0= (0,\dots, 0)$ and $\mu_1= (1/3,\dots, 1/3)$ denote the centers of the majority class and minority class, respectively. Samples for each class are generated from these two distributions, respectively.

We want to study the effect of imbalance between two classes and not focus on the decrease in classification performance due to a decreasing number of observations. Therefore, we fix the number of minority samples $n_1=100$ and increase the number of majority samples $n_0$ in order to change the ratio of the class sizes. With $n_0 \in \{900,1900,9900\}$ we obtain imbalance ratios of $10\%$, $5\%$ and $1\%$.

We split the data into 70\% training and 30\% test data, stratified according to the class indicator. Note that the oversampling techniques may only be applied on the training data. The data generation and the split into training and test data is repeated 100 times. The performance of each classifier is reported as the average AUC and AUPRC and its standard errors over the 100 repetitions.

\begin{figure}[b!]
	\centering
	\includegraphics[width=0.95\textwidth]{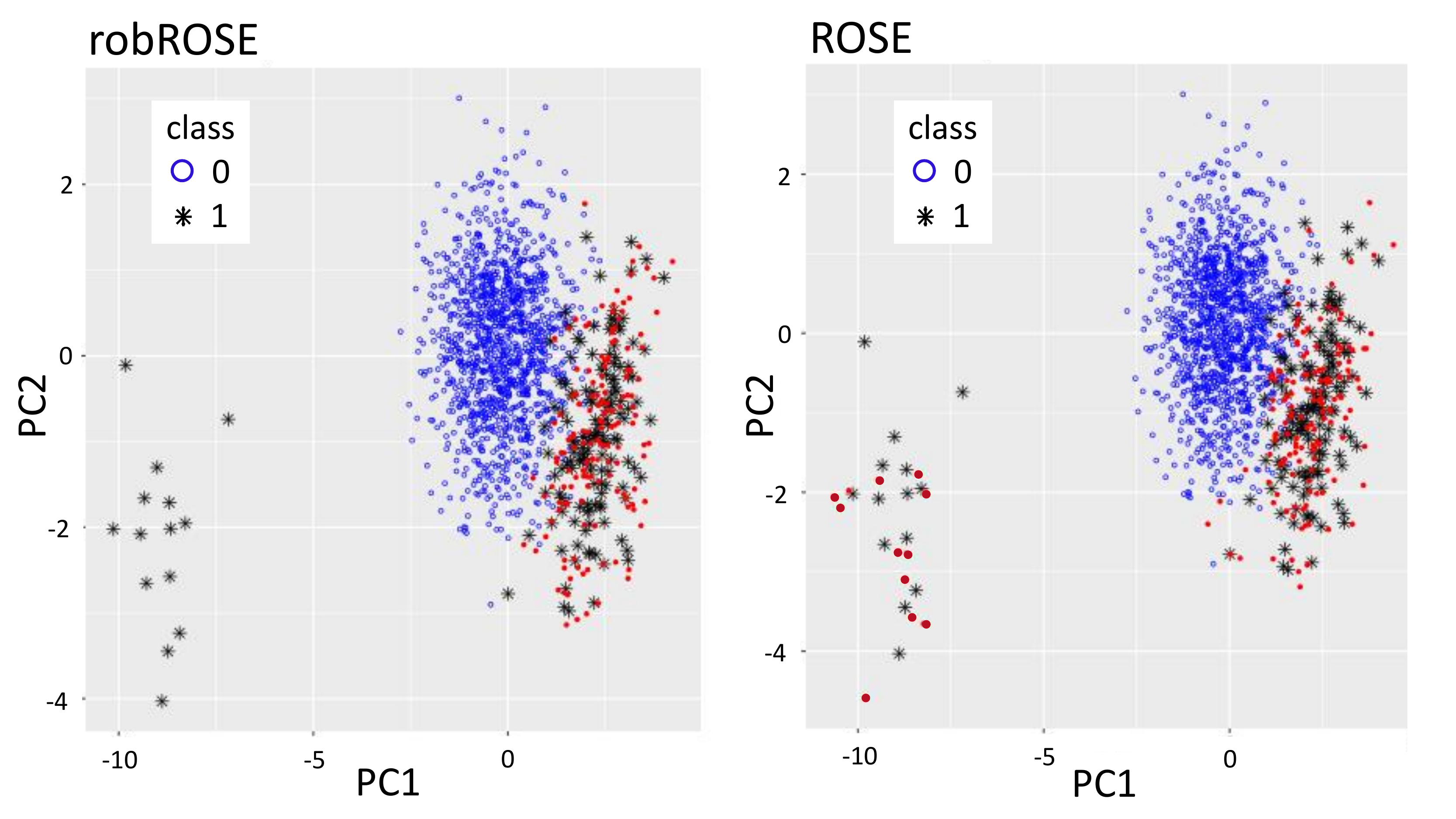}
	\caption{First and second principal component of simulated data (majority class 0 in blue, minority class 1 in black) and artificial samples (in red) generated with robROSE (left) and ROSE (right).}
	\label{fig:pca_oversampling}
\end{figure}
A second simulation setting is used to investigate the effect of anomalous minority samples. In the training set 10\% outliers are generated in the minority class by replacing randomly selected minority samples by samples generated from a multivariate normal distribution with a different center $\mu_{out}=(-10,-2, \dots, -2)$ and the same covariance matrix $\Sigma_1$.

Figure \ref{fig:pca_oversampling} visualizes the first and second principal component of the simulated data with contamination and oversampling by robROSE (left) and ROSE (right).

\subsubsection{Oversampling and model estimation}

In the simulation study we consider four approaches for model estimation with imbalanced data. First, we train the classifiers on the the generated, imbalanced data without applying any balancing strategy before the modeling. Next, we re-balance the training set by using the oversampling methods SMOTE, ROSE and our proposed method robROSE. Finally, we train the classifiers on each of these three oversampled training sets and compare their performance. The oversampling proportion for all settings is 10, i.e. we generate a data set of artificial observations nine times as large as the number of minority samples. The oversampled minority class, which includes both the original and artificial observations, is 10 times the size of the number of original minority samples. On each training data set we estimate a logistic regression model and a robust logistic regression model with the function \texttt{glmrob()} in the \texttt{R} package \texttt{robustbase} \cite{robustbase, cantoni2001robust, valdora2014robust}. Note, however, that the oversampling strategies are model independent.

%%\subsection{Evaluation}
%%AUC and AUPRC with standard error = sd/$\sqrt{nrep}$

\subsection{Simulation results}

\subsubsection{Simulation setting 1 - no outliers}

Results for simulation setting 1 are summarized in Tables \ref{tab:AUPRCsim1} - \ref{tab:AUCsim1rob}. In this setting all oversampling methods (i.e. SMOTE, ROSE and robROSE) slightly improve the results for AUC. For both logistic regression and robust logistic regression all oversampling methods perform equally good in terms of AUC. For AUPRC, there is no evidence of any difference if an oversampling method is applied or not, neither for logistic regression nor for robust logistic regression.

% latex table generated in R 3.5.1 by xtable 1.8-3 package
% Thu Dec 13 18:11:00 2018
\bgroup
\def\arraystretch{1.4}%  1 is the default
\setlength{\tabcolsep}{5pt}
\begin{table}[hbtp]
\begin{center}
\begin{tabular}{rrrrrrrrr}
  \hline
$n_0$ & imbalanced &   & SMOTE &   & ROSE &   & robROSE &   \\ 
  \hline
900 & 0.232 & (.003) & 0.235 & (.003) & 0.235 & (.003) & 0.235 & (.003) \\ 
  1900 & 0.124 & (.002) & 0.125 & (.002) & 0.125 & (.002) & 0.125 & (.002) \\ 
  9900 & 0.027 & (.000) & 0.027 & (.000) & 0.027 & (.000) & 0.027 & (.000) \\ 
   \hline
\end{tabular}
\end{center}
\caption{Simulation setting 1: average AUPRC (and standard error) for logistic regression models (imbalanced refers to no oversampling).}\label{tab:AUPRCsim1}
\end{table}
\egroup

% latex table generated in R 3.5.1 by xtable 1.8-3 package
% Thu Dec 13 18:11:00 2018
\bgroup
\def\arraystretch{1.4}
\setlength{\tabcolsep}{5pt}
\begin{table}[hbtp]
\centering
\begin{tabular}{rrrrrrrrr}
  \hline
$n_0$ & imbalanced &   & SMOTE &   & ROSE &   & robROSE &   \\ 
  \hline
900 & 0.823 & (.003) & 0.835 & (.003) & 0.834 & (.003) & 0.836 & (.003) \\ 
  1900 & 0.812 & (.003) & 0.828 & (.002) & 0.827 & (.002) & 0.829 & (.002) \\ 
  9900 & 0.818 & (.003) & 0.824 & (.002) & 0.824 & (.002) & 0.824 & (.002) \\ 
   \hline
\end{tabular}
\caption{Simulation setting 1: average AUC (and standard error) for logistic regression models.}\label{tab:AUCsim1}
\end{table}
\egroup

% latex table generated in R 3.5.1 by xtable 1.8-3 package
% Thu Dec 13 18:11:00 2018
\bgroup
\def\arraystretch{1.4}
\setlength{\tabcolsep}{5pt}
\begin{table}[hbtp]
\centering
\begin{tabular}{rrrrrrrrr}
  \hline
$n_0$ & imbalanced &   & SMOTE &   & ROSE &   & robROSE &   \\ 
  \hline
900 & 0.227 & (.003) & 0.235 & (.003) & 0.235 & (.003) & 0.234 & (.003) \\ 
  1900 & 0.123 & (.003) & 0.125 & (.001) & 0.125 & (.002) & 0.125 & (.002) \\ 
  9900 & 0.028 & (.001) & 0.027 & (.000) & 0.027 & (.000) & 0.027 & (.000) \\ 
   \hline
\end{tabular}
\caption{Simulation setting 1: average AUPRC (and standard error) for robust logistic regression models.}\label{tab:AUPRCsim1rob}
\end{table}
\egroup

% latex table generated in R 3.5.1 by xtable 1.8-3 package
% Thu Dec 13 18:11:00 2018
\bgroup
\def\arraystretch{1.4}
\setlength{\tabcolsep}{5pt}
\begin{table}[hbtp]
\centering
\begin{tabular}{rrrrrrrrr}
  \hline
$n_0$ & imbalanced &   & SMOTE &   & ROSE &   & robROSE &   \\ 
  \hline
900 & 0.799 & (.003) & 0.837 & (.002) & 0.835 & (.003) & 0.837 & (.002) \\ 
  1900 & 0.780 & (.004) & 0.831 & (.002) & 0.827 & (.002) & 0.831 & (.002) \\ 
  9900 & 0.790 & (.004) & 0.806 & (.003) & 0.803 & (.003) & 0.801 & (.003) \\ 
   \hline
\end{tabular}
\caption{Simulation setting 1: average AUC (and standard error) for robust logistic regression models.}\label{tab:AUCsim1rob}
\end{table}
\egroup

\subsubsection{Simulation setting 2 - with outliers}

Tables \ref{tab:AUPRCsim2} - \ref{tab:AUCsim2rob} present the results for simulation setting 2 where we introduce minority outcasts. Our robROSE method clearly outperforms other oversampling methods in terms of AUPRC and AUC for both logistic regression and robust logistic regression. For logistic regression models using ROSE and SMOTE still improves the results for AUC substantially compared to no oversampling but not for AUPRC.\newline
\indent
Interestingly, robust logistic regression performs worse than logistic regression in the imbalanced setting with outliers in the minority class. The robust method suffers more heavily from the imbalance between both classes than its classical counterpart and has lower AUC and AUPRC. With robust logistic regression, the performance of AUC and AUPRC is improved by all oversampling methods. The best results are obtained by using robROSE which achieves comparable performance for both robust logistic regression and logistic regression models. The performance of the estimators using robROSE is comparable to the performance in the first simulation setting without outliers. We conclude that robROSE give slightly better results than ROSE, although the classifiers also perform well without applying oversampling techniques here. This might be due to the application of principal component analysis as a first preprocessing step, but unfortunately we don't have access to the raw data.

% latex table generated in R 3.5.1 by xtable 1.8-3 package
% Thu Dec 13 18:14:10 2018
\bgroup
\def\arraystretch{1.4}
\setlength{\tabcolsep}{5pt}
\begin{table}[hbtp]
\centering
\begin{tabular}{rrrrrrrrr}
  \hline
 $n_0$ & imbalanced &  & SMOTE & & ROSE &  & robROSE &   \\ 
  \hline
900 & 0.158 & (.003) & 0.156 & (.003) & 0.160 & (.003) & 0.225 & (.003) \\ 
  1900 & 0.084 & (.003) & 0.085 & (.002) & 0.086 & (.003) & 0.125 & (.002) \\ 
  9900 & 0.015 & (.001) & 0.016 & (.001) & 0.016 & (.001) & 0.027 & (.000) \\ 
   \hline
\end{tabular}
\caption{Simulation setting 2: average AUPRC (and standard error) for logistic regression models.}\label{tab:AUPRCsim2}
\end{table}
\egroup

% latex table generated in R 3.5.1 by xtable 1.8-3 package
% Thu Dec 13 18:14:10 2018
\bgroup
\def\arraystretch{1.4}
\setlength{\tabcolsep}{5pt}
\begin{table}[hbtp]
\centering
\begin{tabular}{rrrrrrrrr}
  \hline
 $n_0$& imbalanced &   & SMOTE &   & ROSE &   & robROSE &   \\ 
  \hline
900 & 0.647 & (.005) & 0.679 & (.004) & 0.677 & (.004) & 0.824 & (.003) \\ 
  1900 & 0.624 & (.005) & 0.664 & (.004) & 0.660 & (.005) & 0.828 & (.002) \\ 
  9900 & 0.602 & (.005) & 0.642 & (.004) & 0.643 & (.005) & 0.821 & (.002) \\ 
   \hline
\end{tabular}
\caption{Simulation setting 2: average AUC (and standard error) for logistic regression models.}\label{tab:AUCsim2}
\end{table}
\egroup

% latex table generated in R 3.5.1 by xtable 1.8-3 package
% Thu Dec 13 18:14:10 2018
\bgroup
\def\arraystretch{1.4}
\setlength{\tabcolsep}{5pt}
\begin{table}[hbtp]
\centering
\begin{tabular}{rrrrrrrrr}
  \hline
$n_0$ & imbalanced &   & SMOTE &   & ROSE &   & robROSE &   \\ 
  \hline
900 & 0.153 & (.004) & 0.154 & (.002) & 0.159 & (.003) & 0.231 & (.003) \\ 
  1900 & 0.077 & (.003) & 0.086 & (.003) & 0.087 & (.003) & 0.128 & (.001) \\ 
  9900 & 0.011 & (.000) & 0.016 & (.001) & 0.016 & (.001) & 0.027 & (.001) \\ 
   \hline
\end{tabular}
\caption{Simulation setting 2: average AUPRC (and standard error) for robust logistic regression models.}\label{tab:AUPRCsim2rob}
\end{table}
\egroup

% latex table generated in R 3.5.1 by xtable 1.8-3 package
% Thu Dec 13 18:14:10 2018
\bgroup
\def\arraystretch{1.4}
\setlength{\tabcolsep}{5pt}
\begin{table}[hbtp]
\centering
\begin{tabular}{rrrrrrrrr}
  \hline
$n_0$ & imbalanced &   & SMOTE &   & ROSE &   & robROSE &   \\ 
  \hline
900 & 0.605 & (.006) & 0.682 & (.004) & 0.678 & (.004) & 0.835 & (.002) \\ 
  1900 & 0.564 & (.007) & 0.658 & (.005) & 0.652 & (.005) & 0.836 & (.002) \\ 
  9900 & 0.497 & (.005) & 0.598 & (.006) & 0.601 & (.007) & 0.801 & (.003) \\ 
   \hline
\end{tabular}
\caption{Simulation setting 2: average AUC (and standard error) for robust logistic regression models.}\label{tab:AUCsim2rob}
\end{table}
\egroup

\section{Credit card transaction data}

We consider the Credit Card Transaction Data available at \url{kaggle.com/mlg-ulb/creditcardfraud}. The data consists of transactions made by credit cards in September 2013 by European cardholders. This data set presents transactions that occurred in two days, where we have $492$ frauds out of 284,807 transactions. The data set is highly imbalanced because the positive class (frauds) account for only $0.172\%$ of all transactions. It contains only numerical input variables which are the result of a PCA transformation. Due to confidentiality issues, the original features and more background information about the data is not provided. Features V1, V2, ..., V28 are the principal components obtained with PCA. The only features which have not been transformed with PCA are `Time' and `Amount'. Feature `Time' contains the seconds elapsed between each transaction and the first transaction in the data set. The feature `Amount' is the transaction amount. Feature `Class' is the response variable which takes value $1$ in case of fraud and $0$ otherwise. We select the features V1, V2, ..., V28 and the logarithmically transformed Amount as predictor variables for the classification methods. Each of these 29 predictors are scaled to zero mean and unit variance. To keep the analysis manageable, we only consider main effects and we do not include interactions of any degree.

The following classifiers are used: a decision tree with a maximum depth of 8 built by the CART algorithm \cite{breiman1984classification} and logistic regression. The performance of each classifier is assessed by doing two-fold cross validation 5 times such that in each repetition half of the data is used for trained and the other half is used for testing. The results are summarized in Table \ref{tab:AUC_credit} and \ref{tab:AUPRC_credit}.

\bgroup
\def\arraystretch{1.5}
\setlength{\tabcolsep}{3pt}
\begin{table}[hbtp]
\centering
\begin{tabular}{rrrrrrrrr}
  \hline
 & imbalanced &   & SMOTE &   & ROSE &   & robROSE &   \\ 
  \hline
CART & 0.8990 & (.0181) & 0.9045 & (.0118) & 0.8963 & (.0160) & 0.9102 & (.0118) \\ 
  Logit & 0.9723 & (.0050) & 0.9735 & (.0059) & 0.9766 & (.0046) & 0.9737 & (.0060) \\ 
 % Naive Bayes & 0.9617 & (.0071) & 0.9619 & (.0069) & 0.9601 & (.0077) & 0.9628 & (.0069) \\ 
   \hline
\end{tabular}
\caption{Credit card transaction data: average AUC (and standard error).}\label{tab:AUC_credit}
\end{table}
\egroup

\bgroup
\def\arraystretch{1.5}
\setlength{\tabcolsep}{3pt}
\begin{table}[hbtp]
\centering
\begin{tabular}{rrrrrrrrr}
  \hline
 & imbalanced &   & SMOTE &   & ROSE &   & robROSE &   \\ 
  \hline
CART & 0.7123 & (.0422) & 0.7180 & (.0301) & 0.6428 & (.0581) & 0.7037 & (.0363) \\ 
  Logit & 0.7553 & (.0203) & 0.7622 & (.0210) & 0.7541 & (.0240) & 0.7621 & (.0222) \\ 
%  Naive Bayes & 0.0871 & (.0061) & 0.0894 & (.0065) & 0.0825 & (.0054) & 0.0922 & (.0071) \\ 
   \hline
\end{tabular}
\caption{Credit card transaction data: average AUPRC (and standard error).}\label{tab:AUPRC_credit}
\end{table}
\egroup

\section{Customer churn data}

Imbalanced classification is of course not only a challenge in fraud detection, but is a problem that comes up in many real-world applications. Therefore, classification on imbalanced data sets is a popular topic in machine learning research \cite{krawczyk2016learning}. For example in churn prediction or credit scoring, it is also important to solve the class imbalance problem \cite{zhu2017empirical, marques2013suitability}. 

In this Section we will illustrate the good performance of robROSE on a real churn data set. This data set describes the customer churn of a Korean telecommunication firm focusing only on customers which are companies. Predictor variables are the active months of the customer, its total revenue, the number of employees and the corporate size. These four variables are used to predict customer churn. This is a classic example of imbalanced data since the majority of customers does not churn and we have a small group of minority samples constituting the group of churners. Due to the heterogeneity of companies extreme values can be expected, which may have a heavy influence on the model estimation. This motivates the investigation of oversampling and robust approaches.

The total number of observations in this data set is $n=13601$ (churn: $n_1=3072$, regular: $n_0=10529$). The imbalance ratio is therefore 22.6\% churn which is a rather high proportion for these kind of problems. We reduced the number of churn observations in our study from 22.6\% to 5\% and 1\%, resulting in $n_1\in\{3072,  554,  106 \}$, respectively, to mimic a more extreme imbalance which is more common. 

For each extracted data set the predictor variables were robustly centered by the median and scaled by the median absolute deviation (MAD). Similar as in the simulation framework, 70\% of all observations are used for training and the remaining 30\% is used for assessing the classifier's performance. The training and test set are stratified with respect to the churn indicator such that they have the same class ratio. The split into training and test data is repeated five times.

\bgroup
\def\arraystretch{1.4}
\setlength{\tabcolsep}{5pt}
\begin{table}[hbtp]
\centering
\begin{tabular}{rrrrrrrrr}
  \hline
$n_1$ & imbalanced &   & SMOTE &   & ROSE &   & robROSE &   \\ 
  \hline
3072 & 0.392 & (.008) & 0.358 & (.006) & 0.352 & (.005) & 0.426 & (.002) \\ 
  554 & 0.130 & (.011) & 0.122 & (.012) & 0.124 & (.014) & 0.206 & (.010) \\ 
  106 & 0.055 & (.022) & 0.089 & (.045) & 0.056 & (.017) & 0.124 & (.053) \\ 
   \hline
\end{tabular}
\caption{Churn data Korean corporate: average AUPRC (and standard error) for logistic regression models.}\label{tab:AUPRCchurn}
\end{table}
\egroup

% latex table generated in R 3.5.1 by xtable 1.8-3 package
% Thu Dec 13 18:40:51 2018
\bgroup
\def\arraystretch{1.4}
\setlength{\tabcolsep}{5pt}
\begin{table}[hbtp]
\centering
\begin{tabular}{rrrrrrrrr}
  \hline
$n_1$ & imbalanced &   & SMOTE &   & ROSE &   & robROSE &   \\ 
  \hline
3072 & 0.600 & 0.003 & 0.595 & 0.003 & 0.595 & 0.003 & 0.597 & 0.002 \\ 
  554 & 0.595 & 0.011 & 0.592 & 0.011 & 0.594 & 0.010 & 0.604 & 0.011 \\ 
  106 & 0.590 & 0.046 & 0.596 & 0.051 & 0.598 & 0.050 & 0.609 & 0.050 \\ 
   \hline
\end{tabular}
\caption{Churn data Korean corporate: average AUC (and standard error) for logistic regression models.}\label{tab:AUCchurn}
\end{table}
\egroup

% latex table generated in R 3.5.1 by xtable 1.8-3 package
% Thu Dec 13 18:40:51 2018
\bgroup
\def\arraystretch{1.4}
\setlength{\tabcolsep}{5pt}
\begin{table}[hbtp]
\centering
\begin{tabular}{rrrrrrrrr}
  \hline
$n_1$ & imbalanced &   & SMOTE &   & ROSE &   & robROSE &   \\ 
  \hline
3072 & 0.392 & 0.008 & 0.347 & 0.006 & 0.346 & 0.007 & 0.395 & 0.001 \\ 
  554 & 0.130 & 0.011 & 0.122 & 0.012 & 0.124 & 0.014 & 0.175 & 0.005 \\ 
  106 & 0.055 & 0.022 & 0.087 & 0.044 & 0.075 & 0.027 & 0.109 & 0.052 \\ 
   \hline
\end{tabular}
\caption{Churn data Korean corporate: average AUPRC (and standard error) for robust logistic regression models.}\label{tab:AUPRCchurnrob}
\end{table}
\egroup

% latex table generated in R 3.5.1 by xtable 1.8-3 package
% Thu Dec 13 18:40:51 2018
\bgroup
\def\arraystretch{1.4}
\setlength{\tabcolsep}{5pt}
\begin{table}[hbtp]
\centering
\begin{tabular}{rrrrrrrrr}
  \hline
$n_1$ & imbalanced &   & SMOTE &   & ROSE &   & robROSE &   \\ 
  \hline
3072 & 0.600 & 0.003 & 0.595 & 0.004 & 0.595 & 0.004 & 0.585 & 0.002 \\ 
  554 & 0.595 & 0.011 & 0.592 & 0.011 & 0.596 & 0.011 & 0.598 & 0.009 \\ 
  106 & 0.587 & 0.048 & 0.600 & 0.052 & 0.600 & 0.055 & 0.608 & 0.043 \\ 
   \hline
\end{tabular}
\caption{Churn data Korean corporate: average AUC (and standard error) for robust logistic regression models.}\label{tab:AUCchurnrob}
\end{table}
\egroup

The results are summarized in Tables \ref{tab:AUPRCchurn} - \ref{tab:AUCchurnrob}. The AUC of logistic regression and robust logistic regression is not affected by decreasing the number of churn observations and cannot be improved by any oversampling technique. The AUPRC on the other hand is heavily effected by the increasing imbalance of churn samples. SMOTE and ROSE cannot improve the AUPRC while robROSE achieves substantially better results. The results of AUPRC for robust logistic regression with robROSE are slightly worse than for logistic regression (since the results lie only within one standard deviation).

\section{Conclusion}
Fraud detection can be presented as a binary classification problem with a highly imbalanced class distribution, where fraudsters belong to the minority class. In most applications, the fraud rate is typically less than $0.5\%$. This imbalance problem brings significant challenges to fraud detection and has a great negative impact on standard classification algorithms. Most algorithms tend to bias towards the majority class and may even classify all observations as non-fraudulent, yielding a high overall accuracy but unacceptably low precision with respect to the minority class of interest. 

A popular solution to solve the problem of learning from imbalanced data sets are sampling-based methods. In this paper we focus on synthetic oversampling methods, which have been proven to be very successful in various business applications (e.g. credit scoring, churn prediction and fraud detection).  These methods add new information to the original data set by creating extra synthetic minority class samples based on the existing minority samples that are available in the data set.

In real data sets, it often happens that outliers or anomalies are present. Outliers may be errors, but they could also have been recorded under exceptional circumstances. Outliers can be isolated or may come in clusters, indicating that there are subgroups in the population that behave differently. When outliers are present in the data, the oversampling techniques will generate synthetic samples based on these outliers, which may distort the detection algorithm and make the resulting analysis unreliable. Therefore, it is very important to be able to detect these outliers. 

In practice, one often tries to detect these outliers using traditional diagnostics.
However, classical methods can be affected by outliers so strongly that the resulting fitted model does not allow to detect the deviating observations. This is called the masking effect. 
In the worst case the effect of outliers on a classical fit can be so large that regular observations appear to be outlying, which is known as swamping.
On the other hand, robust methods can resist the effect
of outliers and therefore allow to detect outliers as the observations that deviate substantially from the
robust fit.

In this paper, we have presented a robust version of ROSE, called robROSE, which can cope simultaneously with the problem of imbalanced data and the presence of outliers. Moreover, our robROSE algorithm also offers the advantage that we are able to consider the covariance structure of the minority samples when generating artificial observations. Its good performance is illustrated in a simulation study and on real data sets from fraud detection and churn prediction.

\section*{Acknowledgements}
This work was supported by the BNP Paribas Fortis Chair in Fraud Analytics and Internal Funds KU Leuven under Grant C16/15/068.

\bibliographystyle{plain}
\bibliography{references}

\end{document}